\documentclass[journal]{IEEEtran}
\usepackage{cite}
\usepackage{graphicx}
\usepackage{url}
\usepackage{amsmath}
\usepackage{amssymb}
\usepackage[caption = false, font=footnotesize]{subfig}
\usepackage{float}
\usepackage{booktabs} 

\usepackage{array}
\usepackage{color}
\usepackage{tabularx}
\usepackage{longtable}
\usepackage{tabu}
\usepackage{multirow}
\newcolumntype{L}[1]{>{\raggedright\let\newline\\\arraybackslash\hspace{0pt}}m{#1}}
\newcolumntype{C}[1]{>{\centering\let\newline\\\arraybackslash\hspace{0pt}}m{#1}}
\newcolumntype{R}[1]{>{\raggedleft\let\newline\\\arraybackslash\hspace{0pt}}m{#1}}
\newcommand{\etal}{\textit{et al}.}
\newcommand{\ie}{\textit{i}.\textit{e}.}
\newcommand{\eg}{\textit{e}.\textit{g}.}

\usepackage{bm}
\usepackage{booktabs}
\usepackage[noend]{algpseudocode}
\usepackage{algorithmicx,algorithm}

\usepackage{bbm}

\begin{document}
%
\title{Language-guided Navigation via Cross-Modal Grounding and Alternate Adversarial Learning}

\author{Weixia~Zhang,~\IEEEmembership{Member,~IEEE,}
        Chao~Ma,~\IEEEmembership{Member,~IEEE,}
        Qi~Wu,
        Xiaokang~Yang,~\IEEEmembership{Fellow,~IEEE}
\thanks{This work was supported by National Key Research and Development Program of China (2016YFB1001003), NSFC (61901262, U19B2035, 61527804, 60906119), STCSM (18DZ1112300) and Shanghai Pujiang Program.}
\thanks{W. Zhang, C. Ma and X. Yang are with the MoE Key Lab of Artificial Intelligence, AI Institute, Shanghai Jiao Tong University, China. (e-mail: zwx8981@sjtu.edu.cn; chaoma@sjtu.edu.cn; xkyang@sjtu.edu.cn).}
\thanks{Q. Wu is with the School of Computer Science, The University of Adelaide, Adelaide, Australia (e-mail: qi.wu01@adelaide.edu.au).}
\thanks{{\copyright} 2020 IEEE. Personal use of this material is permitted. Permission from IEEE must be obtained for all other uses, in any current or future media, including reprinting/republishing this material for advertising or promotional purposes, creating new collective works, for resale or redistribution to servers or lists, or reuse of any copyrighted component of this work in other works.}
}


\maketitle

\begin{abstract}
The emerging vision-and-language navigation (VLN) problem aims at learning to navigate an agent to the target location in unseen photo-realistic environments according to the given language instruction. The main challenges of VLN arise mainly from two aspects: first, the agent needs to attend to the meaningful paragraphs of the language instruction corresponding to the dynamically-varying visual environments; second, during the training process, the agent usually imitate the expert demonstrations,\ie, the shortest-path to the target location specified by associated language instructions. Due to the discrepancy of action selection between training and inference, the agent solely on the basis of imitation learning does not perform well. Existing VLN approaches address this issue by sampling the next action from its predicted probability distribution during the training process. This allows the agent to explore diverse routes from the environments, yielding higher success rates. Nevertheless, without being presented with the golden shortest navigation paths during the training process, the agent may arrive at the target location through an unexpected longer route. To overcome these challenges, we design a cross-modal grounding module, which is composed of two complementary attention mechanisms, to equip the agent with a better ability to track the correspondence between the textual and visual modalities. We then propose to recursively alternate the learning schemes of~\textit{imitation} and~\textit{exploration} to narrow the discrepancy between training and inference. We further exploit the advantages of both these two learning schemes via adversarial learning. Extensive experimental results on the Room-to-Room (R2R) benchmark dataset demonstrate that the proposed learning scheme is generalized and complementary to prior arts. Our method performs well against state-of-the-art approaches in terms of effectiveness and efficiency.
\end{abstract}


\IEEEpeerreviewmaketitle
\section{Introduction}
\IEEEPARstart{I}{n} recent years we have witnessed an emergence of navigating an intelligent agent to the target location in realistic \textbf{unseen} environments by natural language, which is of great importance for the agent to interact with a human. The agent needs to understand human natural language instructions, perceive visual surroundings, and take a series of reasonable actions to reach the target location. This emerging vision-and-language navigation task has drawn increasing attention~\cite{wang2018look, fried2018speaker, ma2019self, wang2019reinforced, ma2019regretful, ke2019tactical, landi2019embodied, hu2019you, tan2019learning, huang2019transferable, zhu2020vision, li2019robust} with numerous robotic applications, such as blind guiding, domestic services, goods delivery, medicare, and nursing. While most existing VLN methods pay attention to design non-trivial navigating mechanisms, we focus on exploiting the cross-modal grounding mechanism and a training strategy in this work, which is orthogonal and able to generalize to different VLN models.

We build our model on a cross-modal reasoning agent with two complementary co-grounding modules, namely~\textit{historical co-grounding} and~\textit{mutual co-grounding}. Inheriting the favorable practices of previous arts~\cite{ma2019self,ma2019regretful,wang2019reinforced,ke2019tactical}, we use the historical co-grounding to attend to both textual context and visual context using the decoder hidden states as the query vector. Apart from this, we equip the agent with a mutual co-grounding module to capture the co-attended areas between textual and visual domains, in which both modalities refer to each other using the corresponding aggregated features.

Existing VLN approaches often train the agent by cloning behaviors from expert demonstrations~\cite{bojarski2016end},~\ie, the shortest path from the starting point to the target location. This imitation learning scheme, also termed as~\textit{teacher forcing}~\cite{williams1989learning, peter18vision}, has achieved impressive results on the R2R benchmark dataset~\cite{peter18vision, wang2018look, tan2019learning, zhu2020vision}. However, we observe that the average length of paths a teacher-forced agent takes roughly matches the average length of ground-truth shortest-paths, \eg, 9.48m versus 9.31m on the validation-unseen set of the R2R dataset~\cite{peter18vision} in our implementation, indicating that the teacher-forced agent is likely to early terminate the route whose length is close to the average length on the training set, regardless of how far the agent is away from the target location. Moreover, there exists a discrepancy of action selection between training and inference~\cite{zhang2019bridging}, leading to accumulative errors along the estimated route (\ie, exposure bias~\cite{peter18vision, bengio2015scheduled}).
To handle this issue, instead of imitating expert actions, an alternative solution is sampling actions from the agent's predicted probability distribution at each step during the training process, namely~\textit{student-forcing}~\cite{peter18vision, fried2018speaker, ma2019self, ma2019regretful, ke2019tactical}. Student-forcing does not require learning from the optimal but limited expert demonstrations, allowing the agent to explore different effective routes in the environments by themselves. By doing so, student-forcing helps the agent to address its own mistakes (\eg, the early termination issue) at inference as it is encouraged to explore diverse viable navigation routes from the environments during the training process~\cite{bengio2015scheduled}. Despite the student-forced agents are inclined to achieve higher navigation success rate, they usually yield longer average trajectories as the agent never refer to the shortest paths to the target location during the training process.

\begin{figure*}[t]
    \centering
    \captionsetup{justification=centering}
    \includegraphics[width=1\textwidth]{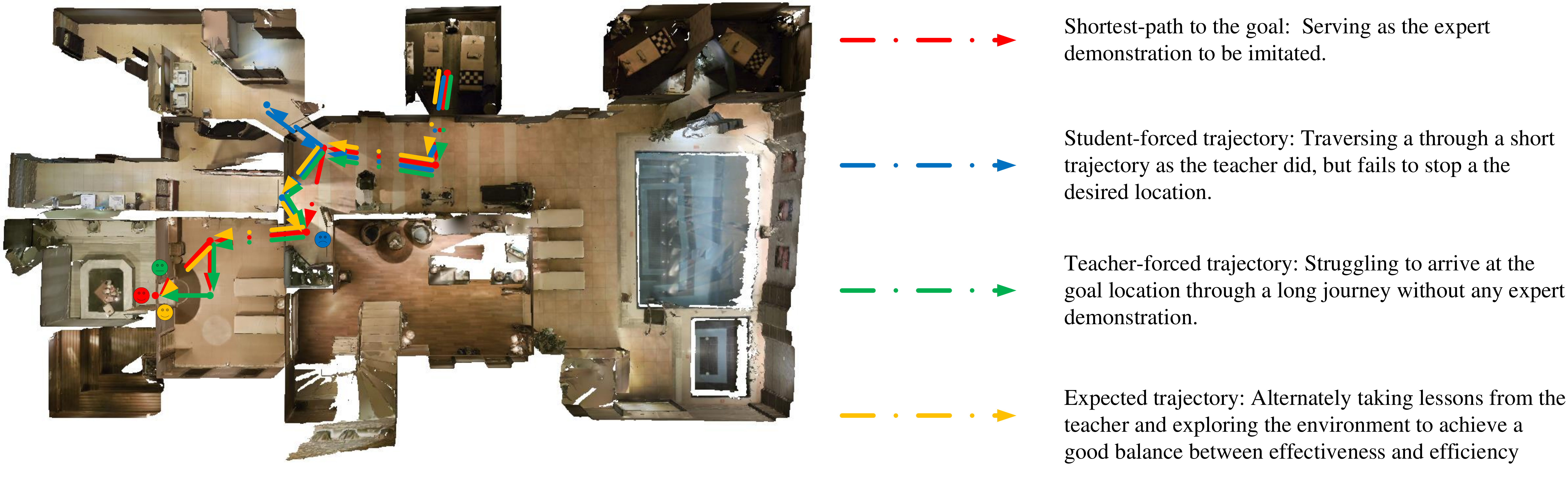}
    \caption{Concept of the motivation. Red dot line: The ground-truth shortest successful trajectory. Green dot line: The trajectory generated by a teacher-forced VLN agent, which is short but not successful. Blue dot line: The trajectory generated by a student-forced VLN agent, which is a successful one but not efficient. Orange dot line: The expected trajectory which is successful and relatively short.}\label{fig:motivation}
\end{figure*}

In this work, we aim to exploit both the advantages of teacher-forcing and student-forcing to train the agent for VLN. To demonstrate our motivation, Figure~\ref{fig:motivation} visualize the ground-truth shortest trajectory, the short but unsuccessful teacher-forced trajectory, the valid but not efficient student-forced trajectory, and the desired trajectory that achieves a good balance between efficiency and effectiveness, respectively. To be specific, we design a simple yet effective learning scheme by \textbf{alternating teacher-forcing and student-forcing} during the training process. Our method not only encourages the agent to imitate the optimal expert demonstrations but also allows it to explore alternative valid navigation routes. In addition, we step further to use adversarial learning to exploit both the advantages of teacher-forcing and student-forcing. Under the adversarial learning context, our language-guided sequence-to-sequence (Seq2Seq) navigator includes a generator of sequential actions and a discriminator that discriminates the different patterns of dynamics of the agent from either teacher-forcing or student-forcing.

Our motivation aligns to \textit{professor forcing}~\cite{lamb2016professor} by jointly considering the advantages with both teacher-forcing (short trajectory length) and student-forcing (high success rate). However, our method differs from professor-forcing in two aspects. First, professor-forcing extends teacher-forcing by encouraging the dynamics of the recurrent network to be the same with that when sampling from the predictions of the model in a unidirectional manner. In contrast, we bidirectionally encourage the dynamics of the recurrent network to be the same as each other during the training process. Second, for implementation, professor-forcing only back propagates gradients from teacher-forced actions, whereas our method alternately back propagates gradients from both the teacher-forced and student-forced actions coupled with the aforementioned~\textit{alternate learning}. The better the agent combines the characteristic of both teacher-forcing (efficiency) and student-forcing (effectiveness), the harder for the discriminator to correctly identify the behavioral sequence of the agent. Therefore, the adversarial learning process is performed as a~\textit{de facto} regularizer to facilitate the advantages combination of teacher-forcing and student-forcing for VLN. This hybrid learning scheme on the basis of both alternate learning and adversarial learning to train the VLN agent is neither used by professor forcing~\cite{lamb2016professor} nor other VLN approaches. As such, we succeed in reducing the discrepancy between training and inference for action selection, leading to a higher level of both the effectiveness and efficiency of the learned navigator.

Overall, our contributions are as follows:
\vspace{0mm}
\begin{itemize}
\setlength\itemsep{0mm}
    \item We construct two complementary co-grounding modules to extract attentive features from textual and visual modalities, leading to large performance gains.
    \item We propose an alternate adversarial learning scheme to mitigate the discrepancy of action selections between training and inference for VLN. Taking both the advantages of student-forcing and teacher-forcing, our alternate adversarial learning is complementary and generalized well to existing VLN arts.
    \item We extensively validate the proposed method on the benchmark R2R dataset. Our method performs well against state-of-the-art approaches in terms of effectiveness and efficiency.
\end{itemize}

\section{Related Work}
\noindent\textbf{Vision and Language Navigation.} Embodied artificial intelligence has drawn increasing attention in recent years~\cite{hemachandra2015learning, vogel2010learning, chen2011learning, zhu2017target, mirowski2017learning, das2018embodied, yu2019multi, CangeaBC19}. Das~\etal~\cite{das2018embodied} formulate an embodied AI task, Embodied Question Answering (EQA), in which an agent is expected to perceive its visual surroundings, take a series of actions to navigate through 3D environments, and answer the questions asked by humans. In this work, we focus on the VLN problem~\cite{peter18vision}. Similar to EQA, a VLN agent is required to navigate through dynamically varying photo-realistic environments to reach a goal location. Instead of answering a question as in EQA, a VLN is driven by a language navigation instruction. Anderson~\etal~\cite{peter18vision} establish a baseline Seq2Seq model with an attention mechanism for the VLN task. Wang~\etal~\cite{wang2018look} introduce model-based and model-free reinforcement learning for this task. Fried~\etal~\cite{fried2018speaker} use data augmentation, pragmatic inference, and panoramic visual representation. The pragmatic inference is not practical in real applications due to extremely long navigation routes~\cite{anderson2018evaluation}, the data augmentation and panoramic visual representation schemes are widely adopted in the following models. Recent representative extensions include monitoring the navigation progress by estimating the normalized distance toward the goal~\cite{ma2019self}, learning to regret and roll back~\cite{ma2019regretful}, searching local optimal navigation routes~\cite{ke2019tactical}, reinforced cross-modal matching and self-supervised imitation learning~\cite{wang2019reinforced}, multi-experts ensemble learning~\cite{hu2019you}, transferable representation learning~\cite{huang2019transferable}. \cite{zhu2020vision} combines a series of auxiliary reasoning tasks~\cite{ma2019self, wang2019reinforced, tan2019learning} to regularize the training process of the VLN agent. An emerging trend is leveraging the power of the large-scale vision-and-language pre-training technology to enhance the representation capability of VLN models~\cite{hao2020towards, majumdar2020improving}. Fusing information from multiple instructions that refer to the same trajectory is also found to boost the performance~\cite{li2019robust, xia2020multi}. In this work, we focus on developing an effective training strategy to address the exposure bias issue along with a cross-modal grounding module to better track the correspondence between the visual and textual domains, which we believe are potentially complementary to the existing work.
According to the used learning schemes, current VLN agents fall into two categories: supervised exploration (student-forcing) approaches~\cite{peter18vision, fried2018speaker, ma2019self, ma2019regretful, ke2019tactical, hu2019you, landi2019embodied}, and collaboration approaches on top of imitation learning (teacher-forcing) and supervised exploration~\cite{li2019robust} or reinforced exploration (reinforcement learning)~\cite{wang2018look, wang2019reinforced, huang2019transferable, tan2019learning, zhu2020vision}. Taking both the efficiency and effectiveness into consideration, in this work, we present a learning approach for VLN by alternating imitation learning and supervised exploration.
For fair comparison to previous state-of-the-art work~\cite{fried2018speaker, ma2019self, wang2019reinforced, ma2019regretful, ke2019tactical, landi2019embodied, tan2019learning, zhu2020vision, li2019robust}, we focus on panoramic action space setting~\cite{landi2019embodied} in this work.\footnote{It is not a requirement of our architecture because we do not make any assumption of the action space.}

\noindent\textbf{Cross-Modal Attention.} The attention scheme is usually effective in bridging the gap between different modalities, such as vision and language. Attention schemes have demonstrated successes on a number of vision-language tasks including image caption~\cite{xu2015show, anderson2018bottom}, visual question answering~\cite{kim2018bilinear, yu2019deep, ma2018visual,wang2017vqa}, visual grounding or referring expression~\cite{deng2018visual, yu2018mattnet,zhuang2018parallel}, etc. In VLN, the widely used soft-attention~\cite{luong2015effective} scheme takes the hidden states of the action decoder as the query vector to obtain attentive textual and visual features. In addition to applying co-attention to textual and visual modalities using the decoder hidden state as the query vector, we further investigate a mutual co-attention module between language instructions and panoramic image features, which facilitates cross-model attentive feature generation at each time step.

\noindent\textbf{Alternate Learning.} Alternating different training patterns periodically is a classical machine learning scheme. A representative example of alternate learning is the Expectation-Maximization (EM) algorithm~\cite{moon1996expectation}, which demonstrates great popularity in various computer vision tasks. Ren~\etal~\cite{ren2015faster} suggest training an object detector,~\textit{i.e.}, Faster-RCNN, by alternately updating the network parameters of the object recognition module and the region proposal module. The most related alternate learning methods to ours are~\textit{scheduled sampling}~\cite{bengio2015scheduled} and DAgger~\cite{ross2011reduction}. Our alternate learning is similar to~\textit{scheduled sampling}~\cite{bengio2015scheduled} in spirit yet differs from it in that our method alternates the learning scheme at each~\textit{episode} with a~\textbf{fixed} curriculum, whereas the scheduled sampling~\textbf{randomly} changes the action decoding rule at each~\textit{step} with some probability. DAgger~\cite{ross2011reduction} trains an initial model with human-annotated data, and then use it to sample trajectories, which will be aggregated with the human-annotated data in a recursive manner. Instead of aggregating the human-annotated and the sampled trajectories to augment the training set as DAgger~\cite{ross2011reduction}, we alternately update our model with either ground-truth data or sampled data and we further exploit adversarial learning to bridge the gap between the dynamics of neural networks with these two learning schemes.

\noindent\textbf{Adversarial Learning.} Goodfellow~\etal introduce a generative framework to synthesize images from random noises with adversarial learning~\cite{goodfellow2014generative}. Training a generative adversarial network (GAN) involves two components,~\textit{i.e.}, a generator for generating images as visually-realistic as possible and a discriminator for distinguishing real images from the synthesized images. The generator and the discriminator evolve simultaneously in iterative competitions between each other. GANs have been used for numerous applications, including conditional image stylization~\cite{isola2017image,zhu2017unpaired}, 
semantic segmentation~\cite{souly2017semi}, object detection~\cite{wang2017fast} and object tracking~\cite{song2018vital}. Adversarial learning also helps the training processes of RNNs for sequential modeling or generation tasks~\cite{lamb2016professor}. In our algorithm, we seamlessly integrate alternate learning into the adversarial learning framework by alternately updating the weights of the generator (navigator) with the gradients from teacher-forced and student-forced actions.

\section{Cross-Modal Navigator}\label{sec:crossmodal_nav}
\subsection{Baseline Navigator}\label{subsec:baseline_navigator}
\begin{figure*}[t]
    \centering
    \captionsetup{justification=centering}
    \includegraphics[width=1\textwidth]{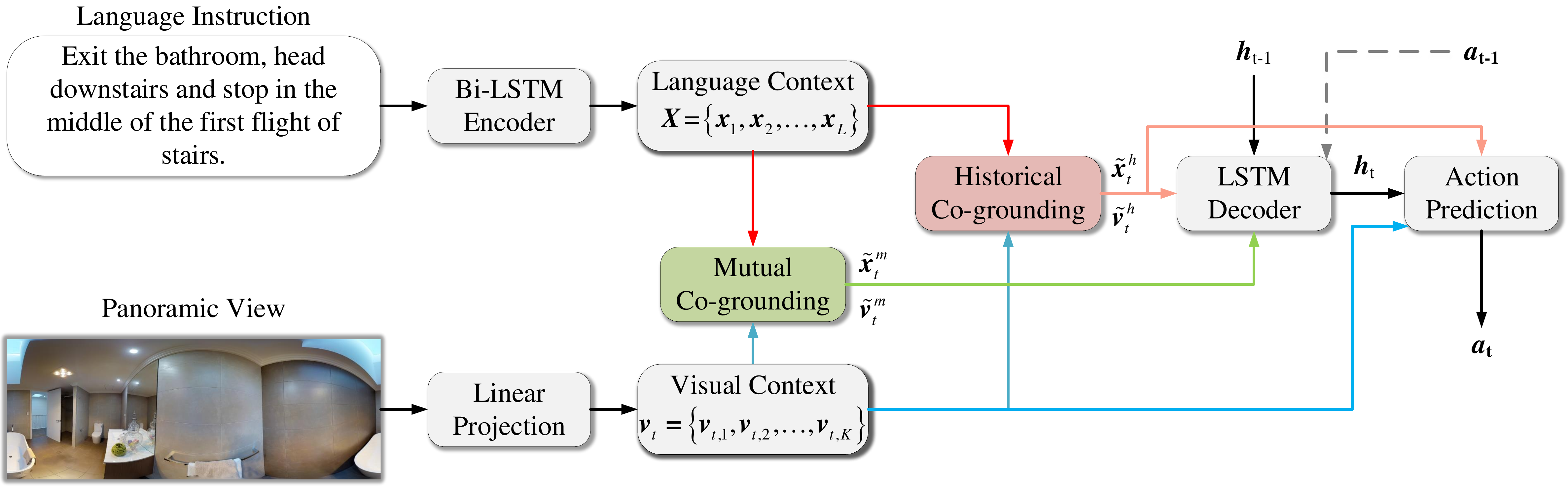}
    \caption{Overview of the baseline navigator, which jointly employs a historical co-grounding and a mutual co-grounding to obtain cross-modal grounded features dynamically. $\bm{\tilde{x}}_{t}^{h}$, $\bm{\tilde{v}}_{t}^{h}$, $\bm{\tilde{x}}_{t}^{m}$, and $\tilde{\bm{v}}_{t}^{m}$ stand for the corresponding grounded features.} \label{fig:baseline navigator}
\end{figure*}

In the VLN task, given a natural language instruction $\bm{I}$ and an initial visual scene which is represented by its image feature $\bm{v}_{0}$, an agent $\pi_{\theta}$ (parameterized by $\theta$) is required to follow the instruction and take a series of actions $\{\bm{a}_{0}, \bm{a}_{1}, ...,\bm{a}_{T}\}$ to move to the desired destination $\bm{v}_{T}$, where $T$ stands for the number of steps executed by the agent in an episode. We use a bi-directional Long Short-Term Memory (Bi-LSTM)~\cite{hochreiter1997long} to encode the instruction into $\bm{X} = \{ \bm{x}_{1}, \bm{x}_{2}, \ldots, \bm{x}_{L}\}$, where $\bm{x}_{l}$ is the encoding of the $l$-th word and $L$ denotes the length of the instruction. Similar to~\cite{fried2018speaker}, we enable the agent to perceive its panoramic visual surrounding. At the $t$-th step, a set of image features $\bm{\upsilon}_{t} = \{ \bm{v}_{t,1}, \bm{v}_{t,2}, \ldots, \bm{v}_{t,K}\}$ is fed to the agent as the panoramic representation of its current viewpoint, where $K$ is the maximum number of navigable directions and $\bm{v}_{t,k}$ denotes the visual representation of direction $k$. We extract the image features using pre-trained ResNet-152~\cite{he2016deep} on ImageNet~\cite{deng2009imagenet} and linearly project them into vectors with a lower dimension using a fully-connected (FC) layer with learnable weights $\bm{W}_{p}$. Figure~\ref{fig:baseline navigator} presents an overview of our baseline navigator.

\subsection{Cross-Modal Grounding Module}\label{subsec:co-grounding}
The agent requires simultaneous grounding of both textual and visual information to identify the ongoing or completed instruction and reason the next direction to go. Different from existing approaches~\cite{ma2019self, ma2019regretful, ke2019tactical, wang2019reinforced}, we propose a cross-modal grounding (CMG) module which is composed of two components,~\textit{i.e.}, a historical co-grounding module, and a mutual co-grounding module.

\subsubsection{Historical co-grounding}\label{subsubsec:his_co}
The hidden states of the decoder, which we implement using an LSTM, track the navigation history. As such, at the time step $t$, we can use the hidden state of the last step of the decoder $\bm{h}_{t-1}$ to get historically grounded visual and textual features, $\bm{\tilde{v}}_{t}^{h} = \sum_{k=1}^{K}\alpha_{t, k}\bm{v_{t, k}}$ and $\bm{\tilde{x}}_{t}^{h} = \sum_{l=1}^{L}\beta_{t, l}\bm{x_{l}}$, respectively, with soft-attentive weights as follows:
\begin{equation}\label{eq:alpha}
\alpha_{t, k} = \mathrm{softmax}(\bm{v}_{t, k}(\bm{W}_{hv}\bm{h}_{t-1})),
\end{equation}
\begin{equation}\label{eq:beta}
\beta_{t, l} = \mathrm{softmax}(\bm{x}_{l}(\bm{W}_{hx}\bm{h}_{t-1})).
\end{equation}
The historical co-grounding is a widely adopted module in existing VLN models~\cite{ma2019self, ma2019regretful, ke2019tactical, wang2019reinforced}. With the dynamically-varying decoder hidden states as the query vectors, the historical co-grounding module endows the agent with the ability to not only identify which parts of a language instruction should be executed but also keep track of the sequence of observed visual information along the navigation trajectory. We inherit this designing philosophy in our model for its effectiveness. Although the historical co-grounding is effective in helping the agent to understand the instruction and perceive the visual surroundings, it relies on the effectiveness of the query vectors, \ie, the decoder hidden states to accurately represent the navigation history. However, a VLN agent may deviate from the shortest path to the goal location and explore other alternative paths during the navigation process, which may degrade the reliability of the decoder hidden states as the query vectors to obtain attended visual and language features.

\subsubsection{Mutual co-grounding}\label{subsubsec:mut_co}
To complement the historical co-grounding module, we propose a mutual co-grounding module which do not relies on the decoder hidden states. Instead of referring to any external signal as the query (as the historical co-grounding), the mutual co-grounding module attends to the most relevant features between textual and visual modalities by referring to each other. Similar to the co-attention mechanism~\cite{ma2018visual}, we first compute a base vector $\mathbf{b}$ by aggregating textual and visual features as follows:
\begin{equation}\label{eq:quer1}
\hat{\bm{v}_{t}} = \mathrm{tanh}(\frac{1}{K}\sum_{k=1}^{K}\bm{v}_{t,k}),
\end{equation}
\begin{equation}\label{eq:query2}
\hat{\bm{x}} = \mathrm{tanh}(\frac{1}{L}\sum_{l=1}^{L}\bm{x}_{l}),
\end{equation}
\begin{equation}\label{eq:query3}
\bm{b}_{t} = \hat{\bm{v}_{t}} \odot \hat{\bm{x}},
\end{equation}
where $\odot$ is the Hadamard product. Then we compute querying vectors for textual grounding and visual grounding, $\bm{q}_{t}^{x}$ and $\bm{q}_{t}^{v}$, respectively, as follows:
\begin{equation}\label{eq:query4}
\bm{q}_{t}^{x} = \mathrm{tanh}({\bm{\tilde{W}}_{x}\hat{\bm{x}}}) \odot \mathrm{tanh}({\bm{W}_{b}\bm{b}_{t}}),
\end{equation}
\begin{equation}\label{eq:query5}
\bm{q}_{t}^{v} = \mathrm{tanh}({\bm{\tilde{W}}_{v}\hat{\bm{v}_{t}}}) \odot \mathrm{tanh}({\bm{W}_{b}\bm{b}_{t}}),
\end{equation}
Based on the query vectors, we can calculate the textual grounded features $\tilde{\bm{x}}_{t}^{m} = \sum_{l=1}^{L}\tilde{\beta}_{t, l}\bm{x_{l}}$ as well as the visual grounded features $\tilde{\bm{v}}_{t}^{m} = \sum_{k=1}^{K}\tilde{\alpha}_{t, k}\bm{v_{t, k}}$ using soft-attention, with attentive weights computed as:
\begin{equation}\label{eq:alpha2}
\tilde{\alpha}_{t, k} = \mathrm{softmax}(\bm{v}_{t, k}(\bm{W}_{qv}\bm{q}_{t-1}^{v})),
\end{equation}
\begin{equation}\label{eq:beta2}
\tilde{\beta}_{t, l} = \mathrm{softmax}(\bm{x}_{l}(\bm{W}_{qx}\bm{q}_{t-1}^{x})),
\end{equation}

\subsubsection{Action prediction}\label{subsubsec:action_predictor}
With all the grounded features, the previous hidden state $\bm{h}_{t-1}$ and the action embedding $\bm{a}_{t-1}$, we obtain the hidden state of the decoder of the time step $t$ as:
\begin{equation}\label{eq:decoder}
\bm{h}_{t} = \mathrm{LSTM}((\bm{\tilde{x}}_{t}^{h} \oplus \bm{\tilde{v}}_{t}^{h} \oplus \tilde{\bm{x}}_{t}^{m} \oplus \tilde{\bm{v}}_{t}^{m} \oplus \bm{a}_{t-1}), \bm{h}_{t-1}),
\end{equation}
where $\oplus$ denotes the concatenation operation.

\noindent Finally, we calculate the probability distribution $\bm{p}_{t}$ over all navigable directions at the time step $t$ as:
\begin{equation}\label{eq:probability1}
\bm{p}_{t} = \mathrm{softmax}(\bm{o}_{t}),
\end{equation}
where the action probability of the $k$-th navigable direction $o_{t, k}$ is computed as the inner-product between the associated image features $\bm{v}_{t,k}$ and the concatenated representation of grounded encoder history context $\bm{\tilde{x}}_{t}^{h}$ along with the decoder hidden state at the time step $t$ as:
\begin{equation}\label{eq:probability2}
o_{t, k} = (\bm{W}_{a}(\bm{h}_{t} \oplus \bm{\tilde{x}}_{t}^{h}))^\top\bm{v}_{t,k}.
\end{equation}
Among the above equations, $\bm{W}_{hv}$, $\bm{W}_{hx}$, $\bm{\tilde{W}}_{x}$, $\bm{\tilde{W}}_{v}$, $\bm{W}_{b}$, $\bm{W}_{qx}$ , $\bm{W}_{qv}$ and $\bm{W}_{a}$ are learnable parameters. We will detail a scheme for selecting the next action $\bm{a}_{t}$ during the training process in Sec.~\ref{sec:AAL}. As for inference, the agent uses greedy decoding to select the next action with the highest predicted action probability.

\section{Alternate Adversarial Learning}\label{sec:AAL}
With the baseline agent equipped with the proposed CMG module stated in Sec.~\ref{sec:crossmodal_nav}, this section presents our training method to encourage the agent to navigate effectively and efficiently. Our motivation is based on the observations from prior arts. Teacher-forced agents constantly imitate the expert demonstrations (shortest paths to the goal) during the training process, likely to finish a single episode with a very short navigation trajectory. Such agents suffer from the problem of exposure bias due to the discrepancy for action selections between training and inference~\cite{lamb2016professor,bengio2015scheduled,li2019robust}, resulting in low navigation success rate. Conversely, a student-forced agent explores the environments and sample actions from its predicted distributions, which is consistent between training and inference, leading to a higher navigation success rate, but at the cost of longer average trajectories. In our scheme, we aim to integrate the strengths of teacher-forcing and student-forcing. We propose two collaborative learning schemes based on the idea of alternate learning and adversarial learning.

\subsection{Alternate Learning}
During the training process, the action at the $t$\textsuperscript{th} step can be either drawn from the ground-truth shortest paths (teacher-forcing) $\bm{a}_{t}^{tf}$ or sampled from the predicted distributions by the resulting model on its own  (student-forcing) $\bm{a}_{t}^{sf}$. We propose to handle the exposure bias by alternating teacher-forcing and student-forcing during the training process, which endows the agent of freedom to explore diverse effective navigation routes and to imitate the golden shortest paths simultaneously. Given the instruction $\{\bm{x}_{i=1}\}_{i=1}^{L}$, we estimate $\theta$ using teacher-forcing by maximizing the log likelihood of ground-truth demonstrations $\{\bm{a}_{t}\}_{t=1}^{T}$ as:
\begin{equation}\label{eq:mle1}
\theta^{*} = \mathop{\arg\max}_{\theta} \mathop\sum \mathrm{log}( \pi_{\theta}(\{\bm{a}_{t}^{tf}\}_{t=1}^{T} | \{\bm{x}_{i=1}\}_{i=1}^{L})).
\end{equation}
When training with student-forcing, we turn to perform maximum log-likelihood estimation of $\theta$ as:
\begin{equation}\label{eq:mle2}
\theta^{*} = \mathop{\arg\max}_{\theta} \mathop\sum \mathrm{log}( \pi_{\theta}(\{\bm{a}_{t}^{sf}\}_{t=1}^{T} | \{\bm{x}_{i=1}\}_{i=1}^{L})).
\end{equation}

\begin{figure*}[t]
    \centering
    \captionsetup{justification=centering}
    \includegraphics[width=1\textwidth]{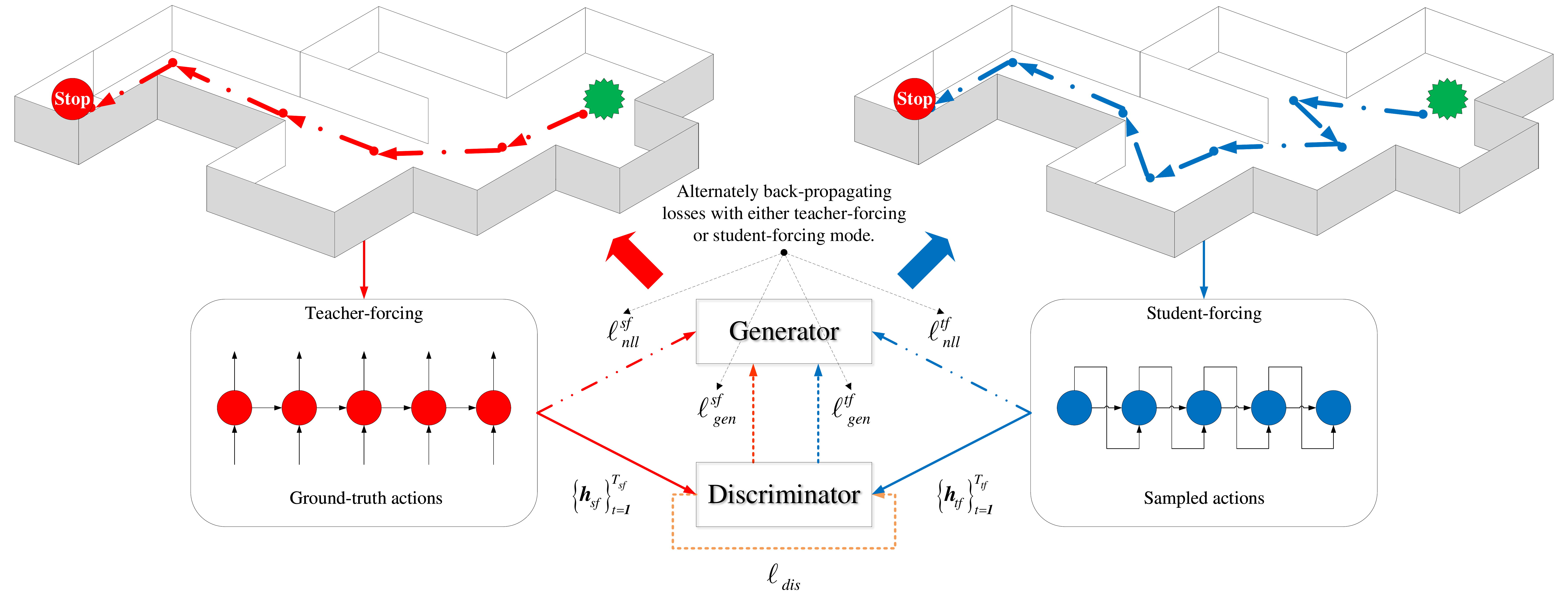}
    \caption{Illustration of our adversarial learning that collaboratively work with the alternate learning. }\label{fig:adversarial}
\end{figure*}

\subsection{Adversarial Learning}
We show that adversarial learning can benefit the agent coupled with both the teacher-forcing and student-forcing learning schemes. Specifically, under the adversarial learning framework, the navigator is treated as a generator $G$ (in terms of generating sequential navigation actions), parameterized by $\bm{\theta}_{g}$. As shown in Figure~\ref{fig:adversarial}, we equip the generator with a discriminator $D$, parameterized by $\bm{\theta}_{d}$. The discriminator $D$ takes as input the behavioral sequence of the generator $G$ to maximize the likelihood of correct classification, \ie, correctly discriminating a behavioral sequence is generated in teacher-forcing or student-forcing mode. Instead of the predicted navigation actions, the behavioral sequence is composed of the decoder hidden states of the generator in the teacher-forcing manner or the student-forcing manner. Therefore, this adversarial learning acts as a regularizer during the training process of the generator $G$. Given an input instruction $\bm{X}$, the output behavioral sequences of the teacher-forced and student-forced generator are denoted by $G_{tf}(\bm{X}, \bm{\theta}_{g})$ and $G_{sf}(\bm{X}, \bm{\theta}_{g})$, respectively. The generator is trained to fool the discriminator by maximizing the probability of $D$ which makes incorrect classifications.

With alternate learning, we perform adversarial learning with two stages. At the first stage, we train the generator with teacher-forcing and backpropagates the negative log-likelihood loss as:
\begin{equation}\label{eq:nll1}
\ell_{nll} = \sum_{t = 1}^{T}-y_{t}\mathrm{log}(\bm{p}_{t}^{tf}),
\end{equation}
where $y_{t}$ is the ground-truth navigable direction at the time step $t$ and $\bm{p}_{t}^{tf}$ is the probability distribution over all navigable directions as previously defined in Eq.~\ref{eq:probability1} where the superscript indicates its running mode. We then update the weights of the discriminator using the standard binary cross entropy (BCE) loss function as:
\begin{align}\label{eq:discriminator_loss1}
\ell_{dis}
&= \sum_{i = 1}^{N}-\mathrm{log}D(G_{tf}(\bm{X}_{i},\bm{\theta}_{g}) , \bm{\theta}_{d}) \nonumber \\
&+ \sum_{i = 1}^{N}-\mathrm{log}(1 - D(G_{sf}(\bm{X}_{i} , \bm{\theta}_{g}) , \bm{\theta}_{d})),
\end{align}
where $N$ stands for the size of the mini-batches randomly sampled from the training data during stochastic gradient descent.

Finally, we optimize the generator to enforce the student-forced behaviors to match teacher-forced behaviors. We label the student-forced behavioral sequences as positive samples with the associated loss as:
\begin{equation}\label{eq:gen_fool1}
\ell_{gen} = \sum_{i = 1}^{N}-\mathrm{log}(D(G_{sf}(\bm{X}_{i} , \bm{\theta}_{g}), \bm{\theta}_{d})).
\end{equation}

In the second stage, we exchange the roles of teacher-forced behaviors and student-forced behaviors. Consequently, all the corresponding loss functions are as follows:
\begin{equation}\label{eq:nll2}
\ell_{nll} = \sum_{t = 1}^{T}-y_{t}\mathrm{log}(\bm{p}_{t}^{sf}),
\end{equation}
and
\begin{align}\label{eq:discriminator_loss2}
\ell_{dis}
&= \sum_{i = 1}^{N}-\mathrm{log}D(G_{sf}(\bm{X}_{i} , \bm{\theta}_{g}) , \bm{\theta}_{d}) \nonumber \\
&+ \sum_{i = 1}^{N}-\mathrm{log}(1 - D(G_{tf}(\bm{X}_{i} , \bm{\theta}_{g}) , \bm{\theta}_{d})),
\end{align}
and
\begin{equation}\label{eq:gen_fool2}
\ell_{gen} = \sum_{i = 1}^{N}-\mathrm{log}(1 - D(G_{tf}(\bm{X}_{i} , \bm{\theta}_{g}), \bm{\theta}_{d})),
\end{equation}
respectively. In this stage, we label the teacher-forced behavioral sequences as the negative samples for the discriminator and thereby encourage the generator to adapt its teacher-forced behaviors $G_{tf}(\bm{X}, \bm{\theta}_{g})$ to be indistinguishable from the student-forced behaviors.

We alternately execute the aforementioned two stages at each iteration during the whole training process. We present the training procedure of our alternate adversarial learning in Algorithm~\ref{algo}.

\begin{algorithm}[t]
\caption{Alternate Adversarial Learning}
\label{algo}
\hspace*{0.02in} {\bf Data:}
Instruction $\{\bm{x}_{l}\}_{l=1}^{L}$, Teacher Actions $\{\bm{a}_{t}^{tf}\}_{t=1}^{T_{tf}}$\\
\hspace*{0.02in} {\bf Models:}
Generator $G$, Discriminator $D$
\begin{algorithmic}[1]
\State initial training mode $m_{1}$ $\leftarrow$ teacher-forcing
\For{i = 1, ..., number of iterations}
    \State ($\{\bm{x}_{l}\}_{l=1}^{L}$, $\{\bm{a}_{t}^{tf}\}_{t=1}^{T_{tf}}$) $\rightarrow$ $G$ $\rightarrow$ ($\{\bm{h}_{tf}\}_{t=1}^{T_{tf}}$, $\ell_{nll}^{tf}$)
    \State ($\{\bm{x}_{l}\}_{l=1}^{L}$, $\{\bm{a}_{t}^{sf}\}_{t=1}^{T_{sf}}$) $\rightarrow$ $G$ $\rightarrow$ ($\{\bm{h}_{sf}\}_{t=1}^{T_{sf}}$,$\ell_{nll}^{sf}$)
    \State ($\{\bm{h}_{tf}\}_{t=1}^{T_{tf}}$, $\{\bm{h}_{sf}\}_{t=1}^{T_{sf}}$) $\rightarrow$ $D$ $\rightarrow$ $\ell_{dis}$
    \If{$m_{i}$ = teacher-forcing}
        \State Update $G$ with $\ell_{nll}^{tf}$
        \State Update $D$ with $\ell_{dis}$
        \State Use $D$ to compute $\ell_{gen}^{tf}$
        \State Update $G$ with $\ell_{gen}^{tf}$
        \State $m_{i+1}$ $\leftarrow$ student-forcing
    \Else
        \State Update $G$ with $\ell_{nll}^{sf}$
        \State Update $D$ with $\ell_{dis}$
        \State Use $D$ to compute $\ell_{gen}^{sf}$
        \State Update $G$ with $\ell_{gen}^{sf}$
        \State $m_{i+1}$ $\leftarrow$ teacher-forcing
    \EndIf
\EndFor
{\bf end for}
\end{algorithmic}
\vspace{0mm}
\end{algorithm}

\section{Experiments}
In this section, we first detail the benchmark dataset and evaluation metrics for performance validation. We then fairly compare our method with state-of-the-art VLN models. In the end, we show the design choices of each module via comprehensive ablation studies.

\subsection{Dataset and Evaluation Metrics}
We evaluate our method on the standard Room-to-Room (R2R)~\cite{peter18vision} vision-and-language navigation dataset, which is built upon the Matterport3D dataset~\cite{chang2017matterport3d}. The R2R dataset contains 7,189 paths captured from diverse navigation graphs in the Matterport3D and 21,567 associated instructions annotated with  Amazon Mechanical Turk (AMT). The dataset is split into training, validation-seen, validation-unseen, and test sets. There are 61 scenes associated with 14,025 instructions being used for training and 1,020 instructions sampled from the same environments are reserved as the validation-seen set. The remaining scenes consist of unseen environments, with 18 scenes associated with 4,173 instructions for the test set and 11 scenes with 2,349 instructions for validation-unseen respectively.

We use three metrics for the evaluation. (1) Navigation Error (NE): the agent's mean navigation error in meters. (2) Success Rate (SR): the agent's mean success rate in terms of reaching the goal location. (3) Success rate weighted by (normalized inverse) Path Length (SPL): as defined in~\cite{anderson2018evaluation}, this metric evaluates the Success Rate against Trajectory Length. Among these three metrics, SR and SPL are primary metrics. We denote by $d(p_{1}, p_{2})$ the shortest distance between two nodes along the edges of the connecting graph in an environment (provided by the simulator). Given $N$ instructions in an evaluation set, NE, SR, and SPL are formulated as Eq.~\ref{eq:NE}, Eq.~\ref{eq:SR}, and Eq.~\ref{eq:SPL}, respectively:
\begin{equation}\label{eq:NE}
NE = \frac{1}{N}\sum_{i = 1}^{N}d(p_{i}, r_{i}),
\end{equation}
where $p_{i}$ and $r_{i}$ refer to the last position of the predicted path and the reference path of the $i$-th navigation instruction, respectively.

\begin{equation}\label{eq:SR}
SR = \frac{1}{N}\sum_{i = 1}^{N}\mathbbm{1}(d(p_{i}, r_{i}) \leq d_{th}),
\end{equation}
where $\mathbbm{1}$ is the indicator function and $d_{th}$ is a distance threshold which is set to 3 meters in the R2R dataset.

\begin{equation}\label{eq:SPL}
SPL = \frac{1}{N}\sum_{i = 1}^{N}\mathbbm{1}(d(p_{i}, r_{i}) \leq d_{th})\frac{l_{i}}{\mathrm{max}(a_{i}, l_{i})},
\end{equation}
where $l_{i}$ and $a_{i}$ denotes the shortest path distance between the starting position to the goal location and the path length actually taken by the agent during executing the $i$-th navigation instruction, respectively.

\subsection{Implementation Details}
In this section, we describe the implementation details of our method. Following previous work, we employ the globally-averaged features extracted from the last convolutional layer of the pre-trained ResNet-152~\cite{he2016deep} on ImageNet~\cite{deng2009imagenet} to represent images. We project the original 2,048-dimensional ResNet-152 features into 512-d vectors with a fully-connected (FC) layer followed by a ReLU unit. We use a variant of environmental dropout~\cite{tan2019learning} with a ratio of 0.4 on image features.\footnote{Different from the original environmental dropout, we only dropout on views and features levels, which helps to achieve better performance in our implementation.}The instruction encoder is composed of a Bi-LSTM with 512-dimensional hidden state (256-d for the forward pass and 256-d for the backward pass) and a 256-dimensional word embedding. We use an LSTM with 512-dimensional hidden states as the decoder to predict actions. As for the discriminator, the hidden states and outputs of the action decoder are summarized by a Bi-LSTM with 512-dimensional hidden states. We use a multi-layer perceptron (MLP) with a sequential structure of~\textit{FC-leaky ReLU-Dropout-FC-leaky ReLU-Dropout-FC-Sigmoid} to output a scalar value~\cite{radford2015unsupervised}, indicating the probability of the input being a teacher-forced or a student-forced behavioral sequence. We augment the training data using a speaker model~\cite{fried2018speaker} trained with the ground-truth trajectory-to-instruction pairs. We use Adam~\cite{kingma2015adam} to perform optimization with a learning rate of 1e-4 and batch size of 64. All the source code will be made available to the public.

\begin{figure*}[t]
    \centering
    \captionsetup{justification=centering}
    \includegraphics[width=1\textwidth]{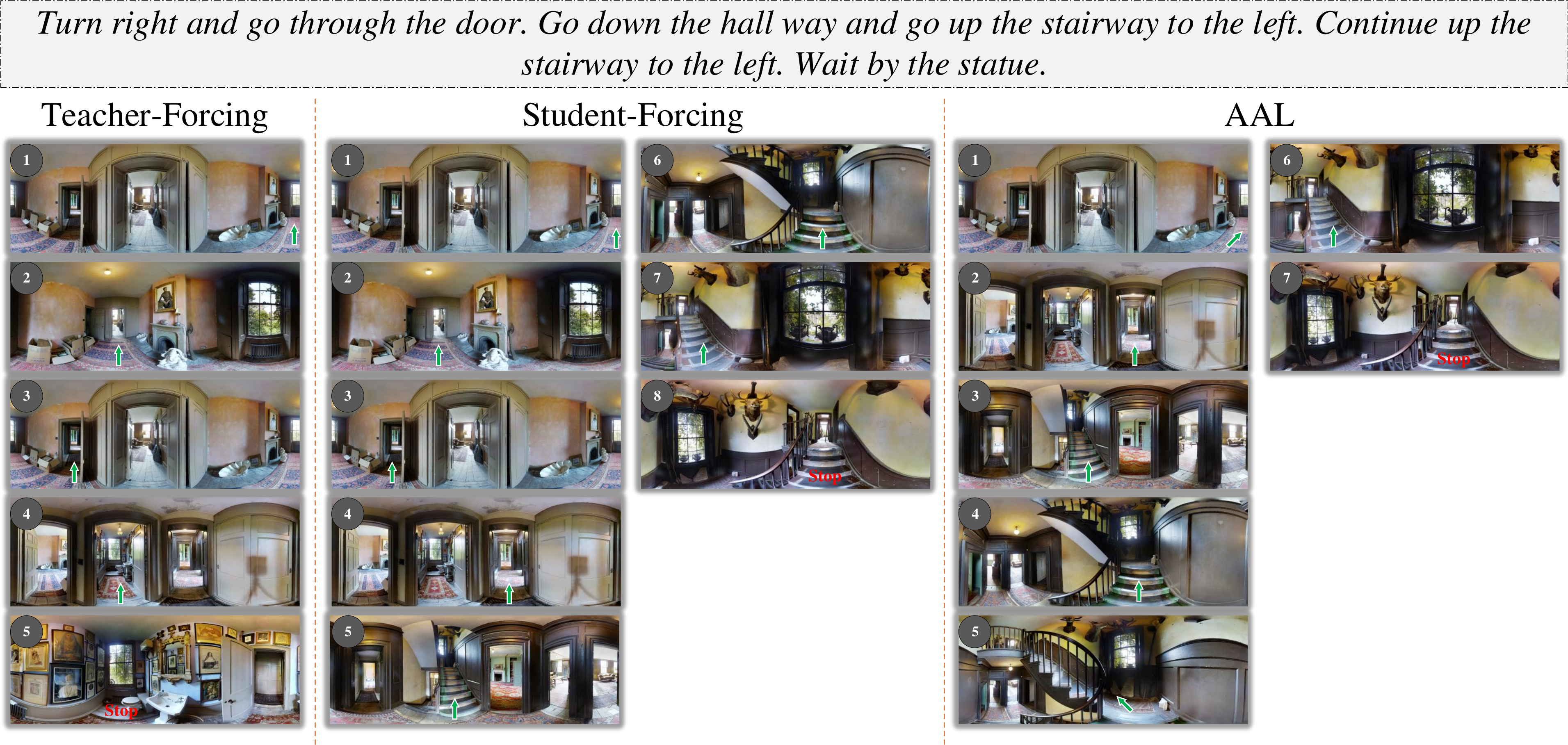}
    \caption{A qualitative result of the proposed CMG-AAL agent, a teacher-forced agent and a student-forced agent. All trajectories correspond to a same instruction.}\label{fig:qualitative1}
\end{figure*}

\begin{figure*}[t]
    \centering
    \captionsetup{justification=centering}
    \includegraphics[width=1\textwidth]{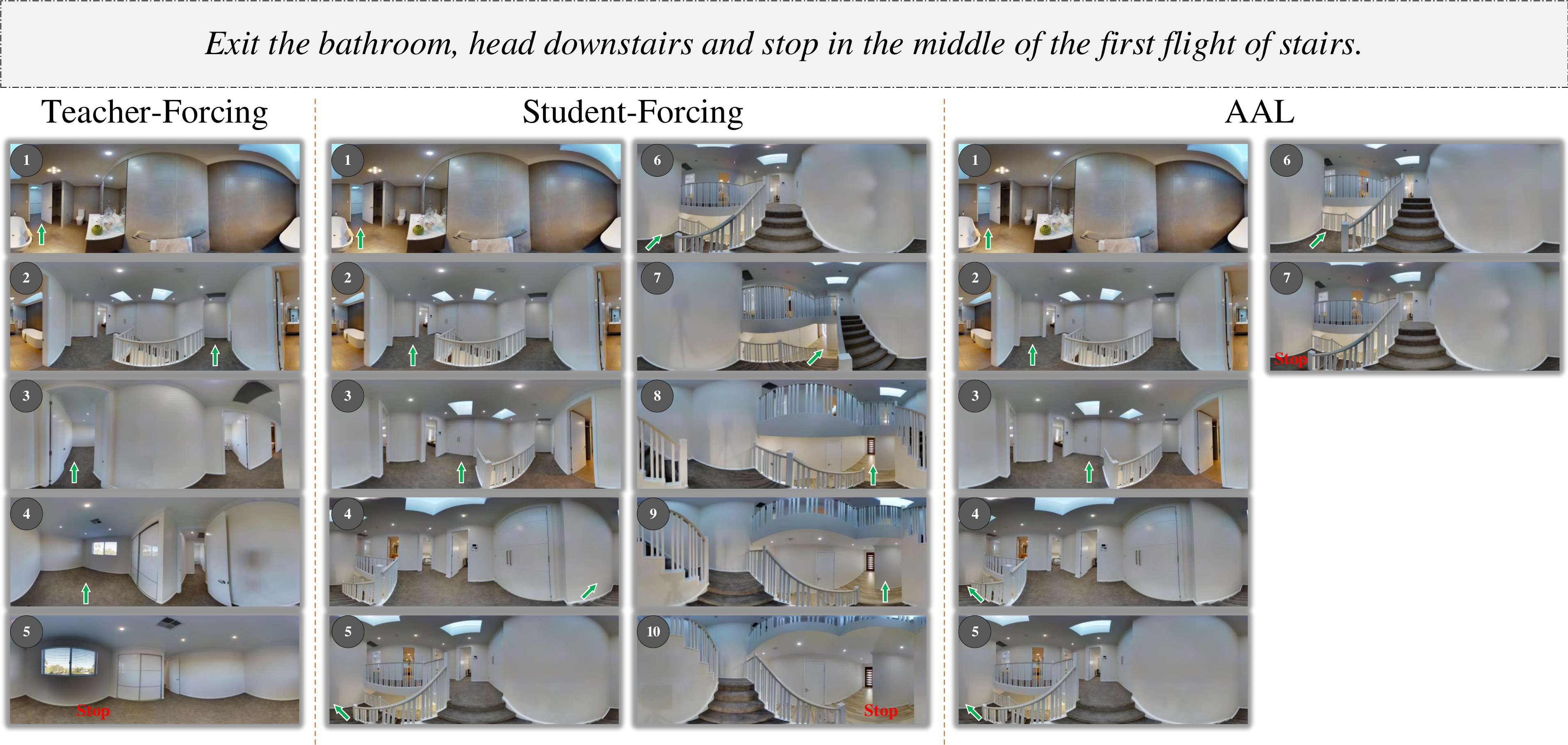}
    \caption{A qualitative result of the proposed CMG-AAL agent, a teacher-forced agent and a student-forced agent. All trajectories correspond to a same instruction.}\label{fig:qualitative1}
\end{figure*}

\begin{figure*}[h]
    \centering
    \captionsetup{justification=centering}
    \includegraphics[width=1\textwidth]{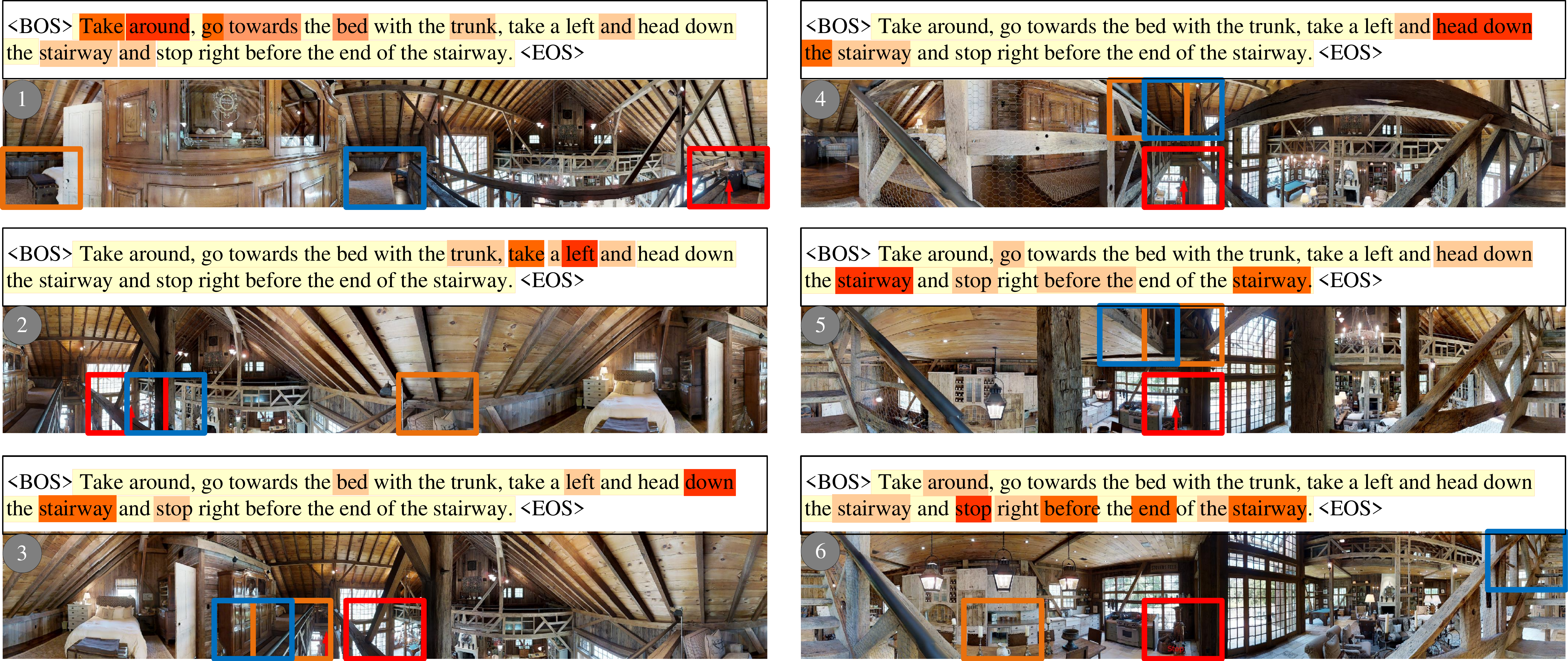}
    \caption{Visual and textual attention visualization during the navigation process. At each step, the textual attention pattern is shown at the top. A deeper color indicates a higher attentive value on the word. The top 3 most attended views in a viewpoint (represented in a panoramic manner) are shown in red, orange, and blue, respectively. The red arrow shows the direction chosen by the agent to move.}\label{fig:attn_pattern}
\end{figure*}

\begin{table*}[t]
  \centering
  \caption{Performance comparison with previous methods. {\bf RL} indicates whether the method use reinforcement learning during the training process or not. {~\bf NE} is navigation error (in meters); lower is better.{~\bf SR} and{~\bf SPL} stand for success rate and success rate weighted by (normalized inverse) path length, respectively; higher is better. $(\ast)$ indicates the results obtained under the single-instruction setting as all other methods conform to.}\label{tab:performance}
  \begin{tabular}{lccccccccccccc}
      \toprule
        &  & \multicolumn{3}{c}{~\bf Validation-Seen} & \multicolumn{3}{c}{~\bf Validation-Unseen} & \multicolumn{3}{c}{~\bf Test ( Unseen )}\\\cmidrule(lr){3-5} \cmidrule(lr){6-8} \cmidrule(lr){9-11}
     {\bf Method} & {\bf RL} & {~\bf NE} $\downarrow$ & {~\bf SR} $\uparrow$ & {~\bf SPL} $\uparrow$ & {~\bf NE} $\downarrow$ & {~\bf SR} $\uparrow$ & {~\bf SPL} $\uparrow$ & {~\bf NE} $\downarrow$ & {~\bf SR} $\uparrow$ & {~\bf SPL} $\uparrow$ \\
     \hline
            Human & & --- & --- & --- & --- & --- & --- & 1.61 & 0.86 & 0.76\\
     \hline
            Seq2Seq~\cite{peter18vision} & & 6.01 & 0.39 & --- & 7.81 & 0.22 & --- & 7.85 & 0.20 & 0.18\\
            RPA~\cite{wang2018look} & \checkmark & 5.56 & 0.43  & --- & 7.65 & 0.25 & --- & 7.53 & 0.25 & 0.23\\
            Speaker-Follower~\cite{fried2018speaker} &  & 3.36 & 0.66 & --- & 6.62 & 0.36  & --- & 6.62 & 0.35 & 0.28\\
            RCM~\cite{wang2019reinforced} & \checkmark & 3.53 & 0.67   & --- & 6.09 & 0.43 & --- & 6.12 & 0.43 & 0.38\\
            Self-Monitoring~\cite{ma2019self} &  & 3.22 & 0.67 & 0.58 & 5.52 & 0.45 & 0.32 & 5.99 & 0.43 & 0.32\\
            Regretful~\cite{ma2019regretful} &  & 3.23 & 0.69 & 0.63 & 5.32 & 0.50 & 0.41 & 5.69 & 0.48 & 0.40\\
            FAST~\cite{ke2019tactical} &   & --- & --- & --- & 4.97 & 0.56 & 0.43 & 5.14 & 0.54 & 0.41\\
            ALTR~\cite{huang2019transferable} &  & 4.68 & 0.56 & 0.53 & 5.61 & 0.46 & 0.43 & 5.49 & 0.48 & 0.45\\
            EnvDrop~\cite{tan2019learning} & \checkmark  & 3.99 & 0.62 & 0.59 & 5.22 & 0.52 & 0.48 & 5.23 & 0.51 & 0.47\\
            PRESS$(\ast)$~\cite{li2019robust} &  & 4.39 & 0.58 & 0.55 & 5.28 & 0.49 & 0.45 & 5.49 & 0.49 & 0.45\\
            AuxRN~\cite{zhu2020vision} & \checkmark  & \bf{3.33} & \bf{0.70} & \bf{0.67} & 5.28 & 0.55 & \bf{0.50} & \bf{5.15} & \bf{0.55} & \bf{0.51}\\
     \hline
            CMG-AAL (Proposed) &  & \bf{2.74} & \bf{0.73} & \bf{0.69} & \bf{4.18} & \bf{0.59} & \bf{0.51} & \bf{4.61} & \bf{0.57} & \bf{0.50}\\
     \bottomrule
   \end{tabular}
\end{table*}

\begin{figure}[h]
    \centering
    \captionsetup{justification=centering}
    \includegraphics[width=0.5\textwidth]{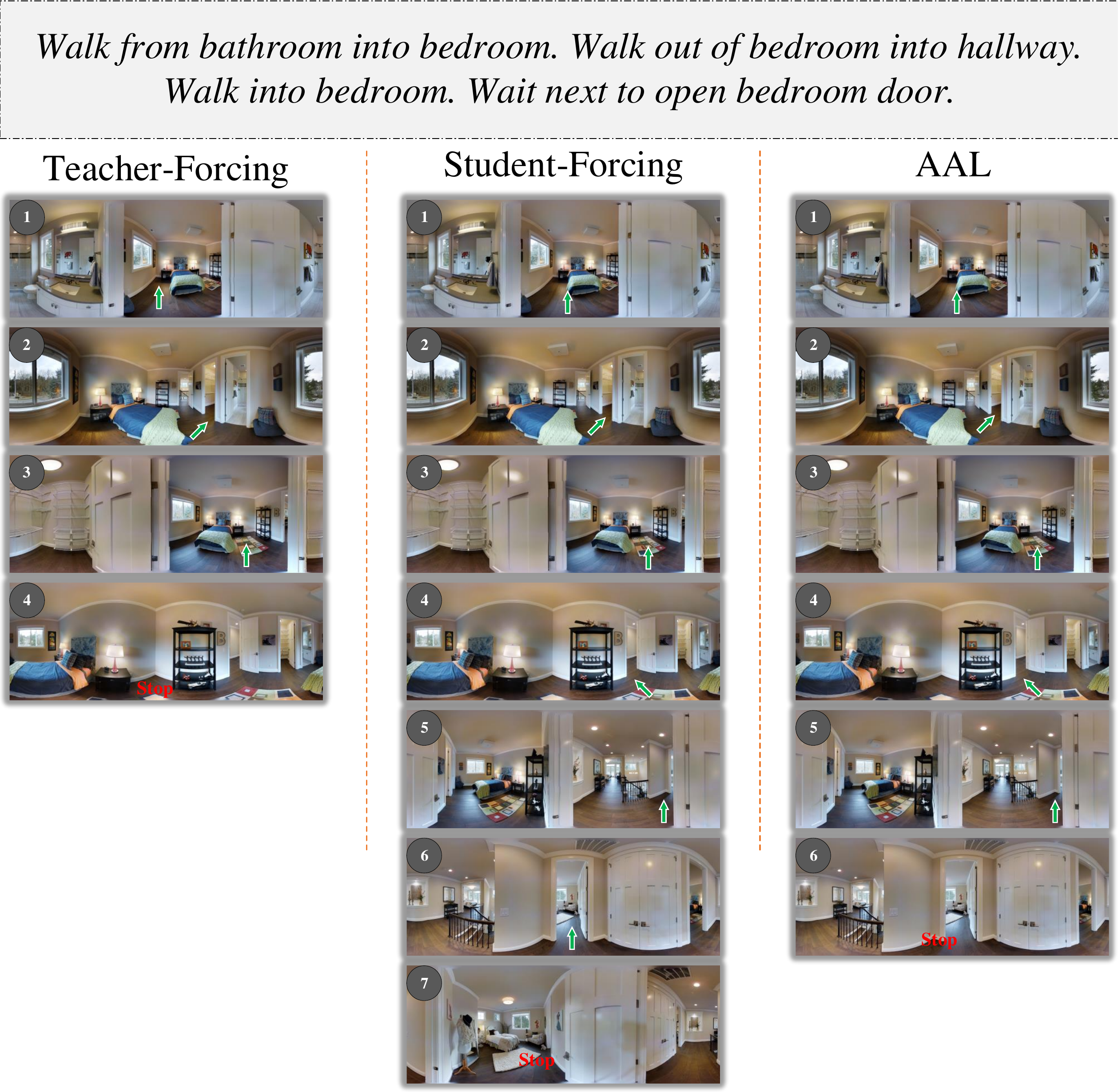}
    \caption{A failure case of the proposed AAL agent, a teacher-forced agent and a student-forced agent. All trajectories correspond to the same instruction. In this example, all of them fail to arrive at the goal location.}\label{fig:fail}
\end{figure}

\subsection{Performance Comparison}
We compare the proposed CMG-AAL agent with state-of-the-art VLN agents. For a fair comparison, we compare with the VLN agents that do not rely on any extra large-scale data for pre-training~\cite{hao2020towards}, which consistently boosts the final performance in terms of all metrics. The experimental results in Table~\ref{tab:performance} show that our method performs favorably against all compared approaches by large margins in terms of two primary metrics SR and SPL. Specifically, we achieve 59\% and 57\% SR on validation-unseen and test sets, respectively, demonstrating 7\% and 6\% performance gains over the reported state-of-the-art results thus far. Compared with the previous state-of-the-art EnvDrop~\cite{tan2019learning} method, our method achieves better SPL results with 3\% absolute improvements on both the validation-unseen and test sets. Note that PRESS~\cite{li2019robust} report higher performance in their original paper under multiple instructions setting,~\ie, aggregating multiple instructions provided for the same trajectories, which consistently boost the final performance in terms of all metrics. We present their results under the single-instruction setting in this table for a fair comparison with other compared methods. By leveraging diverse and complementary supervising signals during the training process, AuxRN~\cite{zhu2020vision} presents competitive performance against our method. While CMG-AAL outperforms AuxRN~\cite{zhu2020vision} in both validation sets in terms of all metrics, AuxRN presents a slightly better performance on the test set in terms of SPL. Another advantage of the proposed CMG-AAL over AuxRN~\cite{zhu2020vision} is that we do not resort to reinforcement learning (which may suffer from different challenges in the real-world scenarios~\cite{dulac2019challenges}) during the training process. The experimental results suggest that the proposed complementary co-grounding modules and the alternate adversarial learning help our agent to take full advantage of both teacher-forcing and student-forcing and deals with the exposure-bias issue well.

For completeness, we also report the experimental results under the pre-exploration~\cite{wang2019reinforced} setting in Table~\ref{tab:pre_explore}. The pre-exploration setting allows the agent to explore the unseen environments before executing navigation. Given the human-annotated instruction, RCM~\cite{wang2019reinforced} pre-explore the unseen environments via a semi-supervised learning method. EnvDrop~\cite{tan2019learning} then uses back-translation to synthesize instruction, with access to the sampled routes from the unseen environments only, leading to a more challenging setting. Similar to~\cite{tan2019learning}, we first pre-train a speaker model~\cite{fried2018speaker} which will be used to generate the synthetic instruction during the pre-exploration process. As shown in Table~\ref{tab:pre_explore}, our method delivers competitive performance against two compared models in terms of NE, SR, and SPL metrics.

\begin{table}[t]
  \centering
  \caption{Performance comparison under the pre-exploration setting.{~\bf TL} denotes trajectory length (in meters).}\label{tab:pre_explore}
  \begin{tabular}{lccccccc}
      \toprule
        & \multicolumn{4}{c}{~\bf Test (Unseen)}\\ \cmidrule(lr){2-5}
        \bf{Method}  & {~\bf NE} $\downarrow$ & {~\bf SR} $\uparrow$ & {~\bf TL} $\downarrow$ & {~\bf SPL} $\uparrow$ \\
     \hline
            RCM~\cite{wang2019reinforced}  & 4.21 & 0.60 & 9.48 & 0.59 \\
            EnvDrop~\cite{tan2019learning}  & 3.97 & 0.64 & 9.79 & 0.61 \\
     \hline
            CMG-AAL  & \bf{3.24} & \bf{0.69} & 11.86 & \bf{0.62} \\
     \bottomrule
   \end{tabular}
\end{table}

\begin{table*}[t]
  \centering
  \caption{Performances of ablation experiments on validation sets.}\label{tab:ablation1}
  \begin{tabular}{lcccccccc}
      \toprule
         & \multicolumn{4}{c}{~\bf Validation-Seen} & \multicolumn{4}{c}{~\bf Validation-Unseen}\\ \cmidrule(lr){2-5} \cmidrule(lr){6-9}
            \bf{Ablated Model} & {~\bf NE} $\downarrow$ & {~\bf SR} $\uparrow$ & {~\bf TL} $\downarrow$ & {~\bf SPL} $\uparrow$ & {~\bf NE} $\downarrow$ & {~\bf SR} $\uparrow$ & {~\bf TL} $\downarrow$ & {~\bf SPL} $\uparrow$\\
     \hline
            Teacher-Forcing & 4.354 & 0.586 & 10.050 & 0.567 & 5.601 & 0.500 & 9.310 & 0.475 \\
            Student-Forcing & 2.666 & 0.722 & 12.361 & 0.648 & 4.535 & 0.562 & 13.562 & 0.469 \\
            Professor-Forcing & 4.490 & 0.586 & 10.027 & 0.565 & 5.670 & 0.490 & 9.155 & 0.467 \\
     \hline
            Historical Co-grounding + Alternate Learning & 3.046 & 0.686 & 11.643 & 0.637 & 4.815 & 0.550 & 12.993 & 0.472 \\
            Mutual Co-grounding + Alternate Learning & 2.926 & 0.703 & 11.593 & 0.651 & 4.664 & 0.558 & 13.162 & 0.475 \\
            CMG + Alternate Learning & 2.954 & 0.703 & 11.859 & 0.651 & 4.362 & 0.576 & 13.105 & 0.491 \\
            Historical Co-grounding + AAL & 2.931 & 0.690 & 11.703 & 0.644 & 4.583 & 0.566 & 13.125 & 0.483 \\
            Mutual Co-grounding + AAL & 2.831 & 0.694 & 11.318 & 0.651 & 4.478 & 0.561 & 12.456 & 0.490 \\
     \hline
            CMG-AAL & \textbf{2.739} & \textbf{0.732} & 11.339 & \textbf{0.687} & \textbf{4.183} & \textbf{0.593} & 12.631 & \textbf{0.514}\\
     \bottomrule
   \end{tabular}
\end{table*}

\subsection{Ablation Studies}
In this section, we conduct a series of ablative experiments on the validation sets to verify the design choice of each component of the proposed CMG-AAL agent. Table~\ref{tab:ablation1} presents the experimental results of all controlled versions of our method. We begin with two baseline agents trained with teacher-forcing and student-forcing, respectively, both of which share the same model structure as described in Sec.~\ref{sec:crossmodal_nav}. As presented in Table~\ref{tab:ablation1}, although the teacher-forced agent can finish navigation with clearly shorter trajectory lengths than the student-forced agent on average, but yielding inferior performance in terms of success rate. These results empirically affirm our assumption that teacher-forcing and student-forcing both have their pros and cons. We then make attempts to exploit the strengths of teacher-forcing and student-forcing using the closely-related learning schemes  professor-forcing~\cite{lamb2016professor}. Interestingly, professor-forcing cannot bring any performance improvements upon the baseline teacher-forcing. Considering that professor-forcing consistently updates the dynamics of RNNs when sampling actions from the network to match the dynamics of RNNs trained with teacher-forcing, it fails to incorporate the advantage of student-forcing into the training process. To demonstrate the necessities of our design choice of each component, we first implement a pure alternate learning method by removing the adversarial learning process. Based on the alternate learning, we train three variants with the historical co-grounding, mutual co-grounding, and the proposed CMG module, respectively. We also train another two models using the proposed AAL method on the navigator equipped with historical co-grounding only and mutual co-grounding only, respectively. The experimental results suggest that the proposed CMG-AAL agent equipped with all the components achieves the best results, suggesting the effectiveness of the cross-modal grounding module and the AAL training strategy.

\begin{table*}[t]
  \centering
  \caption{Performance comparison by integrating AAL to existing approaches on the validation-unseen set. All methods here are trained without data augmentation.}\label{tab:plug-n-play}
  \begin{tabular}{lllll}
      \toprule
            Learning Scheme & {~\bf NE} $\downarrow$ & {~\bf SR} $\uparrow$ & {~\bf TL} $\downarrow$ & {~\bf SPL} $\uparrow$\\
     \hline
            Speaker-Follower & 6.971 & 0.274 & 14.988 & 0.211 \\
            + AAL & 6.541 (\color{blue}-0.430) & 0.337 (\color{blue}+0.063) & 12.539 (\color{blue}-2.449) & 0.294 (\color{blue}+0.083) \\
     \hline
            Self-Monitoring & 5.925 & 0.426 & 15.141 & 0.327 \\
            + AAL & 5.816 (\color{blue}-0.109)  & 0.442 (\color{blue}+0.016) & 14.786 (\color{blue}-0.559) & 0.355 (\color{blue}+0.028) \\
     \bottomrule
   \end{tabular}
\end{table*}

\subsection{Performance with Different Intervals}\label{subsec:interval}
\begin{figure}[h]
    \centering
    \captionsetup{justification=centering}
    \includegraphics[width=0.5\textwidth]{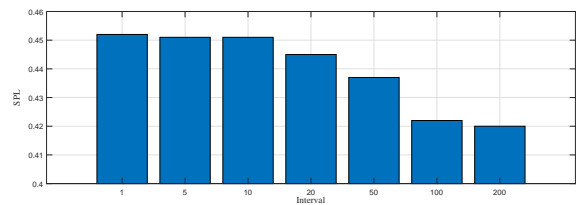}
    \caption{SPL results on the validation-unseen split under different setting of the periodic interval to alternate the training mode between student-forcing and teacher-forcing.}\label{fig:interval}
\end{figure}

As described in Sec.~\ref{sec:AAL}, we alternately switch the training mode between student-forcing and teacher-forcing with a fixed periodic interval. Specifically, we set the interval to 1 in our implementation, \ie, we switch the training mode at each iteration. In this subsection, we conduct an additional experiment on the validation-unseen set to evaluate the performance of our method with different settings of the interval. All the results are obtained with the models trained without data augmentation. We plot the SPL results in Fig.~\ref{fig:interval}. We observe that our method delivers similar performance with small periodic intervals and the SPL results drop with increasing intervals. From these results, we conclude that switching the training mode with a high frequency is the key to take full advantage of two learning schemes. Therefore, we set the interval to 1 in our final implementation.

\subsection{Application to Existing Baselines}
The proposed AAL training strategy is model-agnostic, hence we evaluate the generalizability of our method by integrating it into existing VLN baseline models on the validation-unseen set. Table~\ref{tab:plug-n-play} shows how the AAL method helps Speaker-Follower~\cite{fried2018speaker} and Self-monitoring~\cite{ma2019self} addressing the exposure bias issue.\footnote{Note that our results for both baseline models are different from their original reported numbers, due to differences in software versions, GPU types, and hyper-parameters settings.} We observe that AAL consistently helps both baseline models to achieve higher SR results with lower TL, leading to better performance in terms of the primary metric SPL on the validation-unseen set. In conclusion, the proposed AAL method can be used to enhance the performance of different VLN agents in previously unseen environments in a plug-and-play fashion.

\subsection{Qualitative Results}
To further validate the effectiveness of our model, we qualitatively compare how the proposed agents trained with the proposed alternate adversarial learning, the teacher-forcing, and the student-forcing, navigate through an unseen environment by following the same instruction. As demonstrated in the first example in Figure~\ref{fig:qualitative1}, the trajectories of three agents confirm our previous observations,~\ie, the student-forced agent is inclined to arrive at the desired destination through a relatively long route while the teacher-forced agent is prone to just move a few steps and stop at an undesired location. By comparison, our method successfully advises the agent to exploit a feasible route to reach the goal location through a shorter trajectory than the student-forced agent.

The second example in Figure~\ref{fig:qualitative1} visualizes one more qualitative result. In this example, the teacher-forced agent makes an unwise decision to turn right at the second step. Consequently, it ultimately navigates itself into a room that is severely deviated from the desired direction. An interesting and noteworthy phenomenon is both the student-forced agent and the AAL agent take favorable actions at the first few steps and they both reach the goal at the 5\textsuperscript{th} step. However, instead of stepping on the brakes, both of them decide to keep moving forward. The student-forced agent takes five more steps and ultimately arrives at an undesired location. Conversely, the AAL agent makes a sensible choice to stop at a location that is not much far away from the goal ($<3m$) and is judged as successfully arriving at the target location thereby. It is the use of alternate and adversarial learning to enable the agent to stop taking erroneous actions.

In a more comprehensive perspective, we present a failure case of all the agents trained with the proposed AAL method, the teacher-forcing, and the student-forcing. As shown in Figure~\ref{fig:fail}, all three agents take identical actions in the first three steps and arrive at the same location in the 4\textsuperscript{th} step. The teacher-forced agent then early terminate its trajectory, stopping itself in a place that is far away from the goal location. The student-forced agent and our AAL agent make similar decisions to continue their navigation. Unfortunately, although both of them take two favorable actions to move toward the desired location, they are confused by the two~\textit{open bedroom door} emerging in their panoramic views. While the AAL agent stops itself next to the undesired open bedroom door, the student-forced agent takes one more erroneous action to enter the undesired bedroom. Such a problem can be handled by reducing the ambiguity of language instruction.

To evaluate the effectiveness of the cross-modal co-grounding module, we present the learned visual and textual attention patterns by summing the attentive values of historical and mutual co-grounding modules within a navigation episode in Fig.~\ref{fig:attn_pattern}. In the beginning, the agent focuses on 'take around', followed by 'bed' and 'trunk', which describe the characteristic of the desired location. It then turns to capture the key instruction 'turn a left'. In the following three steps, it successfully keeps its textual attention on the important words 'head down' and 'stairway', navigating itself to go down the stairway. At the end of the navigation, it realizes that it should 'stop right before' the 'end' of the 'stairway'. During the whole navigation process, the most attended views match the final direction the agent decides to go next.

\section{Conclusion}
In this paper, we propose a CMG module and an AAL training strategy for VLN. Being composed of two complementary co-grounding mechanisms, the proposed CMG module helps the agent grounding the dynamic visual environments with the given language instructions. We use episode-wise alternate learning to handle the exposure bias problem caused by the discrepancy of action selection between the training and inference phases. In addition, we propose an adversarial learning method coupled with our alternate learning to take full advantages of two elementary learning schemes for the sequence-to-sequence tasks, namely teacher-forcing and student-forcing, providing an effective and efficient model. We validate the generalization ability of the proposed learning scheme by training prior VLN models using our AAL method, experimental results of which suggest that our method is model-agnostic and help VLN agents perform better in previously unseen environments. Extensive qualitative and quantitative experimental results on the benchmark R2R dataset demonstrate that the proposed approach delivers competitive performance with state-of-the-art approaches.

\bibliographystyle{IEEEtran}
\bibliography{Weixia}

\end{document}